\def\year{2017}\relax
\documentclass[letterpaper]{article}
\usepackage{aaai17}
\usepackage{times}
\usepackage{helvet}
\usepackage{courier}
\usepackage[utf8]{inputenc} 
\usepackage[T1]{fontenc}    
\usepackage{hyperref}       
\usepackage{url}            
\usepackage{booktabs}       
\usepackage{amsfonts}       
\usepackage{nicefrac}       
\usepackage{microtype}      

\usepackage{graphicx}
\usepackage{relsize}
\usepackage{multirow} 
\usepackage{booktabs} 
\usepackage{amsmath} 
\usepackage{amsthm} 
\usepackage{subfigure} 
\usepackage{algorithm}
\usepackage{algorithmic}
\newtheorem*{thm}{Matrix Tree Theorem (MTT)}
\begin{document}

\title{Latent Dependency Forest Models}
\author{Shanbo Chu, Yong Jiang and Kewei Tu\\ 
	School of Information Science and Technology\\
	ShanghaiTech University, Shanghai, China  \\
	$\{$chushb, jiangyong, tukw$\}$@shanghaitech.edu.cn}
\maketitle

\begin{abstract}
	Probabilistic modeling is one of the foundations of modern machine learning and artificial intelligence. In this paper, we propose a novel type of probabilistic models named latent dependency forest models (LDFMs). A LDFM models the dependencies between random variables with a forest structure that can change dynamically based on the variable values. It is therefore capable of modeling context-specific independence. We parameterize a LDFM using a first-order non-projective dependency grammar. Learning LDFMs from data can be formulated purely as a parameter learning problem, and hence the difficult problem of model structure learning is circumvented. Our experimental results show that LDFMs are competitive with existing probabilistic models.
\end{abstract}

\section{Introduction} \label{introduction}
	Probabilistic modeling is one of the foundations of modern machine learning and artificial intelligence, which aims to compactly represent the joint probability distribution of random variables. The most widely used approach for probabilistic modeling is probabilistic graphical models. A probabilistic graphical model represents a probability distribution with a directed or undirected graph. It represents random variables with the nodes in the graph and uses the edges in the graph to encode the probabilistic relationships between random variables. However, traditional probabilistic graphical models have a number of limitations. First, inference takes exponential time in the worst case. Second, learning probabilistic graphical models (in particular, learning the model structures) is very difficult in general. Third, the dependencies between variables are fixed and therefore context-specific independence (CSI) (i.e., independency between variables that only holds given a specific assignment of certain variables) cannot be represented in the basic form of probabilistic graphical models.
	
	To alleviate or solve the first two problems of probabilistic graphical models, a number of tractable probabilistic models are proposed. One example is mixtures of trees (MTs) \cite{meila2001learning}, which represent a probability distribution with a finite mixture of tree distributions. Inference and learning of MTs are both tractable. In addition, certain CSI can be represented in MTs. 
	
	More recently, sum-product networks (SPNs) \cite{poon2011sum} have been proposed as a type of very expressive tractable probabilistic models that subsume many previous probabilistic models as special cases. A SPN can be represented as a rooted directed acyclic graph with univariate distributions as leaves and sums and products as internal nodes.  Exact inference in SPNs can be done in linear time with respect to the network size. However, the large number of latent sum and product nodes in SPNs makes learning (especially structure learning) of SPNs very challenging.
	
\begin{figure}
		\centering
		\includegraphics[width=\columnwidth]{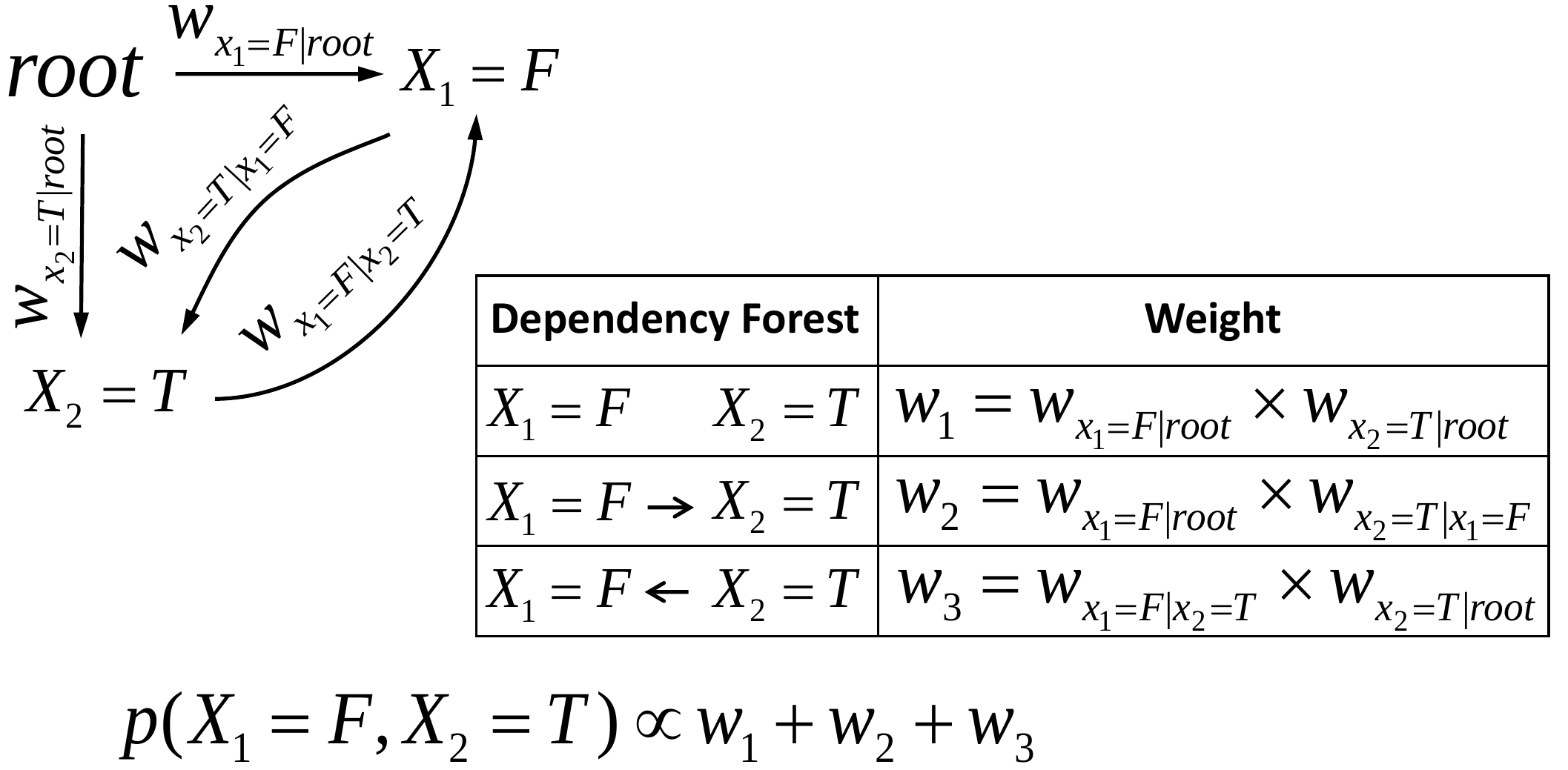}	
		\caption{An example of using LDFMs to compute the joint probability of $X_1 = False$ and $X_2 = True$.  Left: all possible pairwise dependencies between the two variables and a root node (explained in the following section); each dependency has a weight. Right: all three possible dependency forests and their weights. Bottom: computing the joint probability.}		
		\label{an example of LDFMs}
	\end{figure}

	It has been shown that SPNs can be seen as an extension of probabilistic context-free grammars (PCFGs) and a special case of stochastic And-Or grammars \cite{poon2011sum,Tu15}. Based on this observation, we would like to draw a parallel between learning probabilistic models such as SPNs and unsupervised learning of probabilistic grammars. Learning the structure of a probabilistic model resembles learning the set of production rules of a grammar, while learning model parameters resembles learning grammar rule probabilities. From the unsupervised grammar learning literature, one can see that learning approaches based on PCFGs have not been very successful, while the state-of-the-art performance has mostly been achieved based on less expressive models such as dependency grammars (DGs) \cite{klein2004corpus,NAACLHLT09,tu2012unambiguity,SpitkovskyThesis13}. In comparison with PCFGs, DGs eliminate all the latent nodes (i.e., nonterminals) and therefore the difficult discrete optimization problem of structure learning can be easily converted into the more amenable continuous optimization problem of parameter learning.  Inspired by this property of DGs, we propose latent dependency forest models (LDFMs) as a new type of probabilistic models. LDFMs encode the relations between variables with a dependency forest (a set of trees).  A distribution over all possible dependency forests given the current assignment of variables is specified using a first-order non-projective DG \cite{mcdonald2007}. The probability of a complete assignment can then be computed by adding up the weights of all the dependency forests. Figure \ref{an example of LDFMs} gives an example of using LDFMs to compute the joint probability of an assignment of two random variables. The number of possible forests grows exponentially with the number of variables, but the summation of the forest weights can still be computed tractably by utilizing the Matrix Tree Theorem (MTT) \cite{MatrixTree}. 
	
	Compared with existing probabilistic models, LDFMs have the following features. Similar to SPNs, LDFMs model \emph{latent} dependencies between random variables, i.e., the dependencies are dynamically determined based on the assignments of the random variables, and therefore LDFMs are capable of modeling CSI (we give an example of modeling CSI with LDFMs in the supplementary material). Unlike SPNs, there is no latent variable in LDFMs, resulting in easier learning. Similar to MTs, LDFMs assume the latent dependencies to form a forest structure; but unlike MTs that restrict the possible dependency structure to a small number of forest structures, LDFMs consider \emph{all} possible forest structures at the same time. By parameterizing the model using a first-order non-projective DG, unnormalized joint probability computation can still be made tractable. Unlike most of the previous probabilistic models, learning LDFMs from data can be formulated purely as a parameter learning problem without the difficult step of structure learning, and therefore it can be handled by any continuous optimization approach such as the expectation-maximization (EM) algorithm.

\section{Related Work} \label{Related Work}		

\subsection{Probabilistic Modeling}
	
	Probabilistic modeling aims to compactly model a joint distribution over a set of random variables. Many different types of probabilistic models have been proposed in the literature. The most widely used probabilistic models are perhaps Bayesian networks (BNs) \cite{pearl1988bayesian} and Markov networks (MNs) \cite{kindermann1980markov}. A BN models a set of random variables and their conditional dependencies with a directed acyclic graph (DAG). The nodes in the DAG represent the random variables and the edges represent the dependencies. The dependencies are specified between variables regardless of their assignments, so CSI is not representable.  The inference of BNs is generally intractable: computing the exact probability of a marginal or conditional query is $\sharp$P-complete \cite{roth1996hardness}. In addition, learning the structure of BNs from data is very challenging and finding the global optimum structure is known to be NP-hard \cite{chickering2004large}. A MN is similar to a BN in its representation of dependencies, the differences being that BNs are directed and acyclic, whereas MNs are undirected and may be cyclic. In general, exact inference of MNs is also $\sharp$P-complete \cite{roth1996hardness} and learning of MNs is hard.
	
	
	Mixtures of trees (MTs) \cite{meila2001learning} represent joint probability distributions as finite mixtures of tree distributions. One can represent certain CSI in MTs by using different tree distributions to model different contexts. Like most mixture models, the EM algorithm can be used for the learning of MTs. The inference of MTs takes linear time in $n$ (the number of variables) and each step of the EM learning algorithm takes quadratic time in $n$. The number of mixture components in a MT is usually set to a small number in practice. LDFMs can also be seen as a mixture of tree distributions, but unlike MTs, LDFMs consider all possible structures of tree distributions and resort to first-order non-projective DGs to encode the mixture weight of each tree distribution. 
	
	
	A sum-product network (SPN) \cite{poon2011sum} is a tractable deep probabilistic model  represented as a rooted DAG with univariate distributions as leaves and sums and products as internal nodes. A sum node computes the weighted sum of its child nodes and a product node computes the product of its child nodes. The root of a SPN can represent different types of probability distributions according to different inputs. For example, when all the leaves are set to $1$, the root represents the partition function; when the input is evidence, the root represents the unnormalized marginal probability of the evidence; and when the input is a complete assignment, the root represents the unnormalized joint probability. SPNs can also represent CSI by using different sub-SPNs to model distributions of variables under different contexts. Calculating the root value of a SPN is a bottom-up process with time complexity linear in the network size, so inference is tractable in terms of the network size. Structure learning of SPNs is still a challenging problem involving difficult discrete optimization and recently a variety of approaches have been proposed  \cite{gens2013learning,rooshenas2014learning,adellearning}. More recently, some subclasses of SPNs have been proposed \cite{rahman2014cutset,peharz2014learning}.  
	
	For most of the above models, learning involves identification of a good model structure, which is typically a difficult discrete optimization problem. Borrowing ideas from the unsupervised grammar learning literature, we formulate a LDFM as a first-order non-projective dependency ``grammar'' on variable assignments and circumvent the structure learning step by including all possible ``grammar rules'' in the model and then learning their parameters.
	
\subsection{Dependency Grammars} \label{Dependency Grammars}
	
\begin{figure}[t]
			\centering
			\includegraphics[width=\columnwidth]{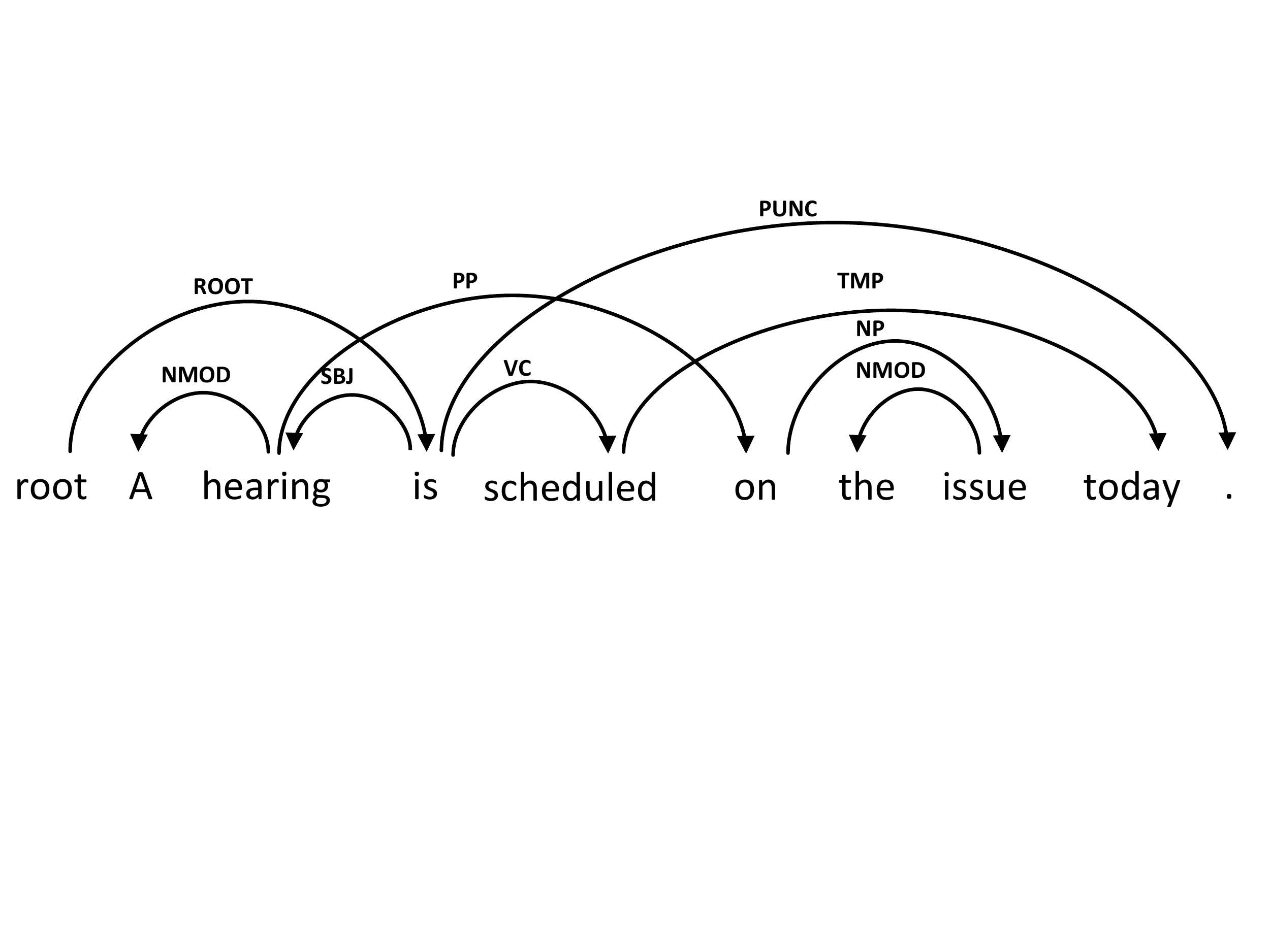}
			\caption{A non-projective dependency tree of an example sentence (from (McDonald and Satta 2007)).}
			\label{non-peojective dependency graph}
	\end{figure}

	In natural language processing (NLP), dependency grammars (DGs) are a simple flexible mechanism for encoding words and their syntactic dependencies through directed graphs. In the directed graph derived from a sentence, each word is a node and the dependency relationships between words are the edges. In a non-projective dependency graph, an edge can cross with other edges. Non-projectivity arises due to long distance dependencies or in languages with flexible order. In most cases, the dependency structure is assumed to be a directed tree. 	Figure \ref{non-peojective dependency graph} is an example non-projective dependency tree for the sentence \emph{A hearing is scheduled on the issue today.} Each edge connects a word to its modifier and is labeled with the specific syntactic function of the dependency, e.g., SBJ for subject.  Dependency parsing (finding the dependency trees) is an important task in NLP and McDonald et al. \cite{mcdonald2005non} proposed an efficient parsing algorithm for first-order non-projective DGs by searching for maximum spanning trees (MSTs) in directed graphs. McDonald and Satta \cite{mcdonald2007} appealed to Matrix Tree Theorem to make unsupervised learning feasible for first-order non-projective DGs via the EM algorithm.
	
	LDFMs can be seen as an extension of first-order non-projective DGs for probabilistic modeling. There are two main differences between non-projective DGs and LDFMs. One difference is that LDFMs use the nodes in a dependency tree to represent the assignments of the variables, while non-projective DGs use the nodes to represent the words in a sentence. The other difference is that when viewed as generative models, a DG always produces a valid sentence while a LDFM may produce invalid assignments of variables if there are conflicting or missing assignments.	

\section{Latent Dependency Forest Models} \label{LDFMs}

\subsection{Basics}
	Let $\mathbf{X}=(X_1,X_2,\ldots,X_n)$ be a set of random variables and $\mathbf{x}=(x_1,x_2,\ldots,x_n)$ be an assignment to the set of random variables.  Given an assignment $\mathbf{x}$, we construct a complete directed graph $G_{\mathbf{x}} = (V_{\mathbf{x}},E_{\mathbf{x}})$ such that,
	\begin{itemize}
		\item $V_{\mathbf{x}} = \{x_0=root, x_1, \ldots, x_n\}$ where $x_0$ is a dummy root node.
		\item $E_{\mathbf{x}} = \{(x_i, x_j) | i \neq j, 0 \leq i \leq n, 1 \leq j \leq n\}$ 
	\end{itemize}
	$G_{\mathbf{x}}$ contains all possible pairwise dependencies between the variables under the current assignments $\mathbf{x}$. We assume that the actual dependency relations between the variables always form a forest structure (a set of trees).
	By adding an edge from the dummy root node to the root of each tree of the dependency forest, we obtain a single dependency tree structure that is a directed spanning tree of the graph $G_{\mathbf{x}}$ rooted at $x_0$. We denote this tree by $T = (V_T,E_T)$, where $V_T = V_{\mathbf{x}}, E_T \subseteq E_{\mathbf{x}}$.		
	
	We assume that the strength of each pairwise dependency is independent of any other dependencies and denote the dependency strength from node $x_i$ to node $x_j$ by edge weight $w_{ij}$. We can compute the strength or weight of a spanning tree $T = (V_T,E_T)$ as the product of the edge weights: 
	\begin{equation*}
	w(T) = \prod_{(x_i,x_j) \in E_T} w_{ij}
	\end{equation*}

	The partition function is the sum over the weights of all possible dependency trees for a given assignment $\mathbf{x}$, which represents the weight of the assignment. We denote this value as $Z_{\mathbf{x}}$.
	
	\begin{equation*}
	Z_{\mathbf{x}} = \sum_{T\in T(G_{\mathbf{x}})} w(T) = \sum_{T\in T(G_{\mathbf{x}})} \prod_{(x_i,x_j) \in E_T} w_{ij}
	\end{equation*}
	$T(G_{\mathbf{x}})$ is the set of all possible dependency trees. The size of $T(G_{\mathbf{x}})$ is exponential in $n$, but we can use Matrix Tree Theorem \cite{MatrixTree} to compute the partition function tractably.

	\begin{thm}
		Let $G$ be a graph with nodes $V = \{x_0,x_1,\ldots,x_n\}$ and edges $E$. Define (Laplacian) matrix $Q$ as a $(n+1)\times(n+1)$ matrix indexed from $0$ to $n$. For all $i$ and $j$, define:
		\begin{equation*}
		Q_{ij} =  
		\begin{cases}
		\mathlarger{\sum_{ i' \neq j, (x_{i'},x_j) \in E}}  w_{i'j} &\mbox{if \  $i=j$}\\
		\ \ \ \ \ \ -w_{ij} &\mbox{if \ $i \neq j$, $(x_i,x_j) \in E$ }
		\end{cases}
		\end{equation*}
		If the $i$-th row and column are removed from $Q$ to produce the matrix $Q^i$, then the sum of the weights of all the directed spanning trees rooted at node $i$ is equal to the determinant of $Q^i$.
		
	\end{thm}
	Thus, to compute $Z_{\mathbf{x}}$, we construct matrix $Q$ from graph $G_{\mathbf{x}}$ and compute the determinant of matrix $Q^0$. The time complexity is $O(n^3)$ if we use LU decomposition for determinant calculation.
	
	Based on the framework introduced above, now we present two generative probabilistic models: LDFM and LDFM-S.

\subsection{LDFM}
	LDFM requires that the weight of each dependency $(x_i, x_j)$ is the conditional probability of generating the variable $X_j$ and setting its value to $x_j$ given the assignment $X_i=x_i$ or the root node. We denote this probability by $w_{x_j|x_i} $. We impose the constraint $0 \leq w_{x_j|x_i} \leq 1$ and the normalization condition  $\sum_{j \neq i} \sum_{x_j} w_{x_j|x_i} = 1 $ for each $x_i$ (or $\sum_{j \neq i} \int_{x_j}w_{x_j|x_i} = 1$ for continuous variables).
	
	An assignment $\mathbf{x}=(x_1,x_2,\ldots,x_n)$ is generated recursively in a top-down manner. First, we generate a dependency tree with $n+1$ nodes uniformly at random. We label the root node as $x_0$. Then, starting from the root node, we recursively traverse the tree in pre-order; at each non-root node, we generate a $\langle variable,value\rangle $ pair conditioned on the $\langle variable,value\rangle $ pair of its parent node.
	The probability of generating an assignment $\mathbf{x}$ is:
	\begin{align*}
	p(\mathbf{x}) = \beta n! Z_{\mathbf{x}} \propto Z_{\mathbf{x}}
	\end{align*}
	where $\beta$ is the uniform probability of the tree structure, and $ \beta n! $ is a constant w.r.t. $\mathbf{x}$. The derivation details can be found in the supplementary material.
	
	
	Note that, however, the above process may also generate invalid assignments. First, it allows multiple assignments of the same variable to be generated at different nodes, resulting in duplicate or conflicting assignments. Second, since there are only $n$ non-root nodes, if some variables are assigned at multiple nodes, then there must be some other variables that are not assigned at all.
	One could modify the generation process to disallow invalid assignments, but that would break the assumed independence between the strength of pairwise dependencies, leading to intractable computation of the partition function $Z_{\mathbf{x}}$.
	
	Since we are only interested in the space of valid assignments (i.e., no duplicate or missing variable assignment), we define the joint probability of a valid assignment $\mathbf{x}$ as \[ \phi(\mathbf{x}) = \frac{p(\mathbf{x})}{\sum_{\mathbf{x} \in A} p(\mathbf{x})} = \frac{Z_{\mathbf{x}}}{\gamma} \]
	where $A$ is the set of valid assignments and $\gamma$ is the normalization factor. The joint probability is proportional to the partition function $Z_{\mathbf{x}}$, so we can use MTT to compute the unnormalized joint probability. However, the normalized joint probability is intractable to compute because computing the normalization factor $\gamma$ is $\sharp$P-hard, which is similar to the case of Markov networks. 
	
	\subsection{LDFM-S}
	The assumption of uniformly distributed tree structures in LDFM can be unreasonable in many cases. In LDFM-S, instead of first uniformly sampling a tree structure and then instantiating the tree nodes, we generate the tree structure and the $\langle variable,value\rangle $ pairs at the tree nodes simultaneously. Starting from the root node, we generate a tree structure in a top-down recursive manner: at each node $x_i$, we keep sampling from the conditional distribution $w_{x_j|x_i}$ to generate new child nodes until a dummy stop node is generated; then for each child of $x_i$, we recursively repeat this procedure.  We denote the probability of generating a stop node given the assignment $X_i=x_i$  by $w_{s|x_i}$ and require the normalization condition $w_{s|x_i} + \sum_{j \neq i} \sum_{x_j} w_{x_j|x_i} = 1$ for each $x_i$ (or $w_{s|x_i} + \sum_{j \neq i} \int_{x_j}w_{x_j|x_i} = 1$ for continuous variables). It is easy to see that if $w_{s|x_i}$ is larger, then $x_i$ is more likely to be a leaf node in the tree structure.
	The probability of generating an assignment $\mathbf{x}$ is:
	{
	\small
	\begin{align*}
	p({\bf x}) =\sum_{T\in T(G_{\mathbf{x}})} \prod_{(x_i,x_j) \in E_T} w_{x_j|x_i} \prod_{x_i \in V_T}  w_{s|x_i} = Z_{\mathbf{x}} \prod_{x_i \in V_{\mathbf{x}}}  w_{s|x_i}
	\end{align*}	
	}

The joint probability of a valid assignment $\mathbf{x}$ is:
\[
\phi(\mathbf{x}) = \frac{p(\mathbf{x})}{\sum_{\mathbf{x} \in A} p(\mathbf{x})} = \frac{Z_{\mathbf{x}} \prod_{x_i \in V_{\mathbf{x}} }  w_{s|x_i}}{\gamma}
\] 
where $A$ is the set of valid assignments and $\gamma$ is the normalization factor. Again, computing the unnormalized joint probability is tractable by using MTT, but computing the normalization factor is $\sharp$P-hard. Note that when all the stop weights $w_{s|x_i}$ are equal, LDFM-S reduces to LDFM.

\section{Inference} \label{inference}
In probabilistic inference, the set of random variables are divided into query, evidence, and hidden variables and we want to compute the conditional probabilities of the query variables given the evidence variables. Inference of LDFMs in general can be shown to be $\sharp$P-hard, so we resort to Markov chain Monte Carlo (MCMC) for approximate inference. 

One way to do MCMC is by Gibbs sampling, which resamples each of the query and hidden variables in turn according to its conditional distribution given the rest of the variables. After getting enough samples, we can compute the probability of a particular query as the fraction of the samples that match the query.
At each resampling step, the conditional distribution is computed from the unnormalized joint probabilities, which takes $O(n^3)$ time and thus can be slow when $n$ is large.

For more efficient sampling, we apply the idea of data augmentation \cite{tanner1987calculation} and simultaneously sample the latent dependency tree structure and the variable values. We name this method tree-augmented sampling. We first randomly initialize the values of the query and hidden variables as well as the dependency tree structure. At each MCMC step, we randomly pick a variable and simultaneously change its value and its parent node in the dependency tree (with the constraint that no loop is formed). Suppose variable $X_i$ is picked, then the proposal probability of value $x_i$ and parent $X_j$ is proportional to the dependency weight $w_{x_i|x_j}$ where $x_j$ is the value of $X_j$ in the previous sample.
It can be shown that the acceptance rate of the proposal is always one. The time complexity of each sampling step is $O(n)$, which can be further reduced if the dependencies between variables are sparse. 
After getting enough samples, we estimate the query probability based on the statistics of the sampled variable values while disregarding the sampled dependency trees.

\setlength{\tabcolsep}{5pt}
\begin{table*}[t]
	\small
	\caption{Dataset Statistics. \emph{Dimension} represents the number of variables. \emph{Avg cardinality} represents the average number of values a variable can take and \emph{Max cardinality} represents the maximum number of values a variable can take. \emph{Avg degree} represents the average number of edges connected to each node in the ground truth BN and \emph{Max in-degree} represents the max number of inward edges directed to a node in the ground truth BN.}
	\label{table:dataset}
	\centering
		\begin{tabular}{|l|l|l|l|l|l|l|l|l|l|}
			\hline
			Dataset & Asia & Child & Alarm  & Sachs & Insurance &  Water & 	Win95pts & Hepar2 & Hailfinder \\ 
			\hline
			Dimension & 8 & 20 & 37 & 11 & 27 & 32 & 76 & 70 & 56 \\
			\hline
			Avg cardinality & 2 & 3 & 2.84 & 3 & 3.30 & 3.63 & 2 & 2.31 & 3.98  \\
			\hline		
			Max cardinality & 2 & 6 & 4 & 3 & 5 & 4 & 2 & 4 & 11 \\
			\hline	
			Avg degree & 2 & 2.5 & 2.49 & 3.09 & 3.85 & 4.12 & 2.95 & 3.51 & 2.36\\
			\hline
			Max in-degree & 2 & 2 & 4 & 3 & 3 & 5 & 7 & 6 & 4\\
			\hline 				
		\end{tabular}
\end{table*}

\section{Learning} \label{Learning}
	We want to learn a LFDM form data where the dependency structure of each training instance is unknown. We avoid the difficult structure learning problem by assuming a complete LDFM structure, i.e., we assume that all the dependencies between $\langle variable,value\rangle $ pairs are possible (having nonzero weights), rather than trying to identify a subset of possible dependencies. We then rely on parameter learning to specify the weights of all the dependencies. This strategy is quite different from that of learning other types of probabilistic models, as structure learning is unlikely to circumvent for most of them. For BNs, if we assume a complete structure (i.e., the skeleton being a complete graph), then the model size (in particular, sizes of conditional probability tables) becomes exponential in the number of random variables and learning becomes intractable. For SPNs, there is no general principle of constructing a ``complete'' structure; if we assume that there is at least one node for each possible scope (subset of random variables), then the model size is also exponential. 

	Our learning objective function is the log-likelihood of the model.
\begin{align*}
	\sum_{\alpha = 1}^{|D|} \log p(\mathbf{x}_{\alpha}) = \sum_{\alpha = 1}^{|D|} \log Z_{{\mathbf{x}}_{\alpha}} + \mathrm{constant}
\end{align*}
	where $D = \{\mathbf{x}_{\alpha}\}_{\alpha = 1}^{|D|} $ is the training dataset. Note that we compute the likelihood based on $p(\mathbf{x})$, the probability of generating an assignment, instead of $\phi(\mathbf{x})$, the probability of a valid assignment. This makes our learning algorithm tractable and also encourages the learned model to be more likely to produce valid assignments.

	Since the dependency structure of each training sample is hidden, we can use the expectation-maximization (EM) algorithm for learning LDFM parameters. We uniformly initialize all the dependency weights, which we find to be the most effective initialization method.


	In the E-step, we compute the partition functions and edge expectations of the training samples. For a training sample $\mathbf{x}$, the edge expectation of $(x_i,x_j)$ is defined as follows: 
\begin{equation*}
	\langle (x_i,x_j) \rangle_{\mathbf{x}} = \sum_{T \in T(G_{\mathbf{x}})} w(T) \times I((x_i,x_j),T)
\end{equation*}
	where $ I((x_i,x_j),T)$ is an indicator function which is one when $(x_i,x_j)$ is in the tree $T$ and zero otherwise. Following the work of McDonald and Satta \cite{mcdonald2007}, we can compute the edge expectations through matrix inversion. 
	When $i,j > 0$,
\begin{equation*}
	\langle (x_i,x_j) \rangle_{\mathbf{x}} =  w_{x_j|x_i} Z_{\mathbf{x}}  [((Q^0)^{-1})_{jj} -((Q^0)^{-1})_{ji}]
\end{equation*}
	When $i = 0$ and $j > 0$,
\begin{equation*}
	\langle (x_0,x_j) \rangle_{\mathbf{x}} = w_{x_j|x_0} Z_{\mathbf{x}}  ((Q^0)^{-1})_{jj}
\end{equation*}

In the M-step, we update the parameters $w_{x_j|x_i}$ to maximize the log-likelihood subject to the constraints of $\sum_{j \neq i}\sum_{x_j} w_{x_j|x_i} = 1 $ (for discrete variables) and $w_{x_j|x_i} \geq 0$. By solving the above constrained optimization problem with the Lagrange multiplier method, we can get:
\begin{align*}
w_{x_j|x_i} =  \frac{\sum_{\alpha = 1}^{|D|}\frac{1}{Z_{{\mathbf{x}}_{\alpha}}} \langle (x_i,x_j) \rangle_{\mathbf{x}_{\alpha}}}{\sum_{\alpha = 1}^{|D|}\frac{1}{Z_{\mathbf{x}_{\alpha}}} \sum_{j' \neq i} \sum_{x_{j'}}  \langle (x_i,x_{j'}) \rangle_{\mathbf{x}_{\alpha}}}
\end{align*}
For continuous variables, we can assume a certain form of conditional distributions (e.g., a bivariate conditional normal distribution) and derive its optimal parameters in terms of the partition functions and edge expectations in a similar manner. 

In addition to maximum likelihood estimaton, we may also run maximum a posteriori (MAP) estimation using EM with a prior over the parameters. We find that the modified Dirichlet prior \cite{MdirTu2016}, which is a strong sparsity prior, can sometimes significantly improve the learning results.


\begin{table*}[t]
	\small
	\caption{The maximum of CLL and CMLL normalized by the  number of query variables. The bold numbers mark the best performance. g-LDFM denotes the results of LDFM by using Gibbs sampling and t-LDFM denotes the results of LDFM by using tree-augmented sampling. }
	\label{results}
	\vskip 0.15in
	\begin{center}
		\begin{tabular}{|@{}c@{}|@{}c|c|c|c|c|c|@{}c|c|c|c|c|c@{}|}
			\hline
			5000 training instances & \multicolumn{6}{|c|}{40\% Query, 30\% Evidence } & \multicolumn{6}{|c|}{30\% Query, 20\% Evidence } \\
			\hline
			Dataset & g-LDFM & t-LDFM & BN & DN & SPN & MT & g-LDFM & t-LDFM & BN & DN & SPN & MT \\ \hline 
			Asia & -0.258 & {\bf -0.241} & -0.274 & -0.268 & -0.262 & -0.262 & -0.268 & {\bf -0.240} & -0.286  & -0.276 & -0.272& -0.272 \\
			Child & {\bf -0.609} &  -0.663 & -0.721 & -0.634 & -0.630 & -0.707 & {\bf -0.650} & -0.702 & -0.744 & -0.670 & -0.688 &-0.761 \\ 
			Alarm & -0.293 & -0.357 & -0.436 & -0.317 & {\bf -0.277} & -0.343 & -0.335 & -0.396 & -0.473 & -0.375 & {\bf -0.328} & -0.379 \\
			Insurance & {\bf -0.460 } & -0.538 & -0.565 & -0.499 & -0.476  & -0.557 & {\bf -0.550} & -0.616 & -0.660 & -0.598 & -0.581 & -0.652 \\
			Sachs & {\bf -0.605} & -0.620 & -0.675 & -0.610 & -0.644 & -0.647 & -0.632 & {\bf -0.627} & -0.698 & -0.649 & -0.683 & -0.681 \\
			Water & {\bf -0.399} & -0.401 & -0.474 & -0.407 & -0.415 & -0.435 & {\bf -0.434} & -0.439 & -0.496 & -0.445 & -0.457 & -0.478\\
			Win95pts & -0.166 & -0.169 & -0.229 & -0.185 & {\bf -0.118} & -0.121  & -0.163 & -0.193 & -0.251 &-0.207 & -0.147 & {\bf -0.146} \\
			Hepar2 & {\bf -0.481} & -0.483 & -0.509 & -0.490 & -0.489 & -0.507 & {\bf -0.481} & -0.482 & -0.513 & -0.493 & -0.497 & -0.513\\
			Hailfinder & -0.991 &  -1.091 & -1.223 & -1.089 & {\bf -0.941} & -1.241 & -1.013 & -1.096  & -1.242 & -1.110 & {\bf -0.979} & -1.268\\
			\hline 
			
			 500 training instances & \multicolumn{6}{|c|}{40\% Query, 30\% Evidence } & \multicolumn{6}{|c|}{30\% Query, 20\% Evidence } \\
			\hline
			
			Dataset & g-LDFM & t-LDFM & BN & DN & SPN & MT & g-LDFM & t-LDFM & BN & DN & SPN & MT \\ \hline 
			Asia & -0.263 & {\bf -0.246} & -0.301 & -0.266 & -0.272 & -0.264 & -0.268 & {\bf -0.244} & -0.296 & -0.280 & -0.276 & -0.273\\
			Child & {\bf -0.623} & -0.677  & -0.801 & -0.668 & -0.757 & -0.927 & {\bf -0.665} & -0.706 & -0.813 & -0.701 &-0.804 & -0.898\\
			Alarm & {\bf-0.328} & -0.368 & -0.521 & -0.359 & -0.426 & -0.526 & {\bf -0.370} & -0.403 & -0.605 & -0.408 & -0.463 & -0.510\\
			Insurance & {\bf -0.478} & -0.542 & -0.665 & -0.533 & -0.596 & -0.706 & {\bf -0.564} & -0.617 & -0.751 & -0.621 & -0.698 & -0.776\\
			Sachs & {\bf -0.627} & -0.634 & -0.702 & -0.653 & -0.759 & -0.733 & {\bf -0.644} & -0.654 & -0.712 & -0.678 & -0.780 & -0.723\\
			Water & {\bf -0.406} & -0.411 & -0.482 & -0.431 & -0.511 & -0.531 & {\bf -0.441} & -0.445 & -0.507 & -0.462 & -0.541 & -0.543\\
			Win95pts & -0.162 & -0.191 & -0.271 & -0.205 & {\bf -0.151} & -0.174 & -0.188 & -0.205 & -0.311 & -0.231 & {\bf -0.173} & -0.194\\
			Hepar2 & -0.491 & {\bf -0.488} & -0.539 & -0.503 & -0.531 & -0.641 & -0.490 & {\bf -0.485} & -0.535 & -0.504 & -0.535 & -0.591\\
			Hailfinder & {\bf -1.024} & -1.10 & -1.449 & -1.127 & -1.187 & -2.244 & {\bf -1.040} & -1.098 & -1.511 & -1.135 & -1.197 & -1.938\\
			\hline
		\end{tabular}
	\end{center}
\end{table*}

\section{Experiments} \label{Experiments}

We empirically evaluated the learning and inference of LDFMs on nine datasets and compared the performance against several  popular probabilistic models including Bayesian networks (BNs), dependency networks (DNs), mixtures of trees (MTs), and sum-product networks (SPNs). 

To produce our training and test data, we picked nine BNs that are frequently used in the BN learning literature from bnlearn\footnote{http://www.bnlearn.com/bnrepository/}, a popular BN repository. For each BN, we sampled two training sets of 5000 and 500 instances, one validation set of 1000 instances, and one testing set of 1000 instances. All the random variables are discrete. Table \ref{table:dataset} shows the statistics of each BN. It can be seen that many of the ground truth BNs are much more complicated than a tree structure and some have relatively high tree-width. Note that the way we produced our data actually gives BNs an advantage in our evaluation because the data never goes beyond the representational power of BNs. Nevertheless, as we will see, LDFMs and a few other models still outperform BNs on these datasets.

We learned the five types of models from the training data and evaluated them by their accuracy in query answering on the test data. For each test data sample, we randomly divided the variables into three subsets: query variables $Q$, evidence variables $E$, and hidden variables $H$. We then ran the learned model to compute the conditional probability $p(Q=q|E=e)$, where $q$ and $e$ are the values that $Q$ and $E$ take in the test data sample. A model that is closer to the ground truth would produce a higher conditional probability on average. Four different proportions of dividing the query, evidence, and hidden variables (e.g., $30\%$ query variables, $10\%$ evidence variables, and $60\%$ hidden variables) were used and for each proportion, a thousand query instances were generated from the test set. Following the evaluation metrics of the previous work \cite{rooshenas2014learning}, we report the maximum of the conditional log-likelihood (CLL) and the conditional marginal log-likelihood (CMLL): $\sum_{X_i \in Q} \log p(X_i = x_i|E = e)$. We normalize CLL or CMLL by the number of query variables to facilitate comparison across different query proportions. 

We trained LDFM using EM and modified Dirichlet prior on each dataset and used the two MCMC approaches introduced in the
\emph{Inference} section to estimate the query probabilities. 
For the other four probabilistic models, we used the Libra toolkit \cite{libra} to train them and tuned the hyperparameters of the training algorithms according to the query probabilities on the validation set. Specifically, Libra learns BNs with decision-tree CPDs \cite{chickering1997bayesian} which is an extension of plain BNs that is capable of modeling CSI; it learns DNs using an algorithm similar to \cite{heckerman2001dependency}; it uses the algorithm proposed in \cite{meila2001learning} to learn MTs; and it learns SPNs using direct and indirect variable interactions \cite{rooshenas2014learning}. Note that in Libra, the input data to the MTs and SPNs training algorithms needs to be binary, so we binarized all the variables when training MTs and SPNs. After training the four probabilistic models, we again used Libra to do inference. For BNs and DNs, we used the Gibbs sampling algorithm implemented in Libra. For MTs and SPNs, Libra first converted the trained models to an equivalent arithmetic circuit (AC) and then used an exact AC inference method to do inference. 

We report the evaluation results of two proportions in Table \ref{results}, and report the results of the other two proportions in the supplementary material. The evaluation results of LDFM-S are similar to the results of LDFM, and we report them in the supplementary material. It can be seen that LDFMs are competitive with the other probabilistic models and achieve the best results on most datasets. Comparing the performance of the two MCMC approaches of LDFMs, we see that Gibbs sampling achieves better overall results than tree-augmented sampling; however, in our experiments tree-augmented sampling is more than twenty times faster than Gibbs sampling on average. SPNs have very good performance on the larger training set 
, which verifies their effectiveness in probabilistic modeling compared with traditional approaches; however, their performance is not as good on the smaller training set 
suggesting that SPNs may require more data to learn than LDFMs. BNs perform worst on average even though the data was generated from ground truth BNs, which suggests that structure learning of BNs is still a very challenging problem.


\section{Conclusion} \label{conclusion}
In this paper, we propose latent dependency forest models (LDFMs), a novel probabilistic model. A LDFM models the dependencies between random variables with a forest structure that can change dynamically based on the variable values. 
We define a LDFM as a generative model parameterized by a first-order non-projective DG. We propose two MCMC approaches, Gibbs sampling and tree-augmented sampling, for inference of LDFMs. Learning LDFMs from data can be formulated purely as a parameter learning problem, and hence the difficult problem of model structure learning is circumvented. We derive an EM algorithm for learning LDFMs. Our experimental results show that LDFMs are competitive with existing probabilistic models.


\section{Acknowledgments}
This work was supported by the National Natural Science Foundation of China (61503248).

\section{Supplementary Material}
	
\subsection{An Example of Using LDFMs to Model CSI}

The assignments of the variables can influence the distribution over the dependency structures. In this way, LDFMs can model CSI to some extent. Here is an example of using LDFM-S to model three binary variables $X_1$, $X_2$ and $X_3$.

\begin{figure}[th]
	\centering
	\includegraphics[width=\columnwidth]{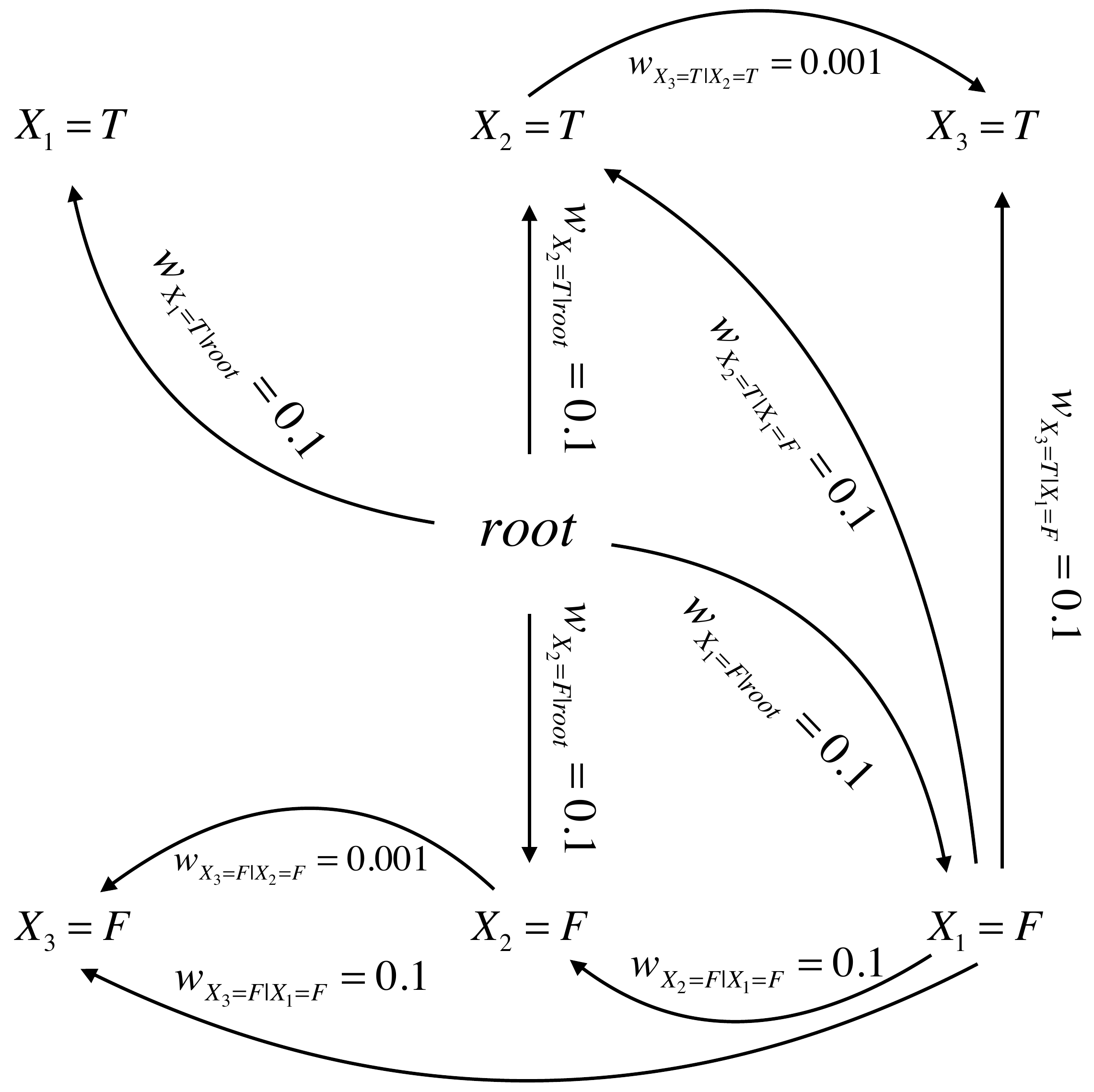}
	\caption{All possible pairwise dependencies between the three variables and a root node. Each dependency has a weight and only the dependencies with non-zero weights are shown. The weight $w_{s|x_i}$, which is the probability of generating a stop node given the assignment $X_i=x_i$ is not drawn for simplicity, but it can be computed using the normalization condition discussed in the \emph{LDFM-S} subsection in the main text.}
	\label{pro}
\end{figure}

Figure \ref{pro} gives an example of using LDFM-S to model CSI.  The conditional probabilities of the two variables $X_2$ and $X_3$ given $X_1$ can be computed using the formula in the \emph{LDFM-S} subsection and they are  shown in Table \ref{condTable}. It can be seen that when $X_1=T$, $X_2$ and $X_3$ are strongly dependent; when $X_1=F$, they are only weakly dependent.

\begin{table}[t]
	\small
	\caption{The conditional probabilities of the two variables $X_2$ and $X_3$ given $X_1$}
	\label{condTable}
	\centering
	\begin{tabular}{|c|c|c|c|}
		\hline
		$X_1$ & $X_2$ & $X_3$ & $P(X_2,X_3|X_1)$ \\
		\hline
		\hline
		T  &  T  &  T & 0.5 \\
		T  &  T  &  F &  0 \\
		T  &  F  &  T &  0 \\
		T  &  F  &  F & 0.5 \\
		F  &  T  &  T & 0.251 \\ 
		F  &  T  &  F & 0.249 \\ 
		F  &  F  &  T & 0.249 \\
		F  &  F  &  F & 0.251 \\
		\hline
	\end{tabular}
\end{table}

\begin{table*}[t]
	\small
	\caption{ The maximum of CLL and CMLL normalized by the  number of query variables. The bold numbers mark the best performance.}
	\label{ldfms}
	\centering
	\begin{tabular}{|l|c|c|c|c|c|c|c|c|c|}
		\hline
		Dataset & Asia & Child & Alarm & Insurance  & Sachs& Water  & Win95pts & Hepar2  & Hailfinder \\
		
		\hline
		\hline
		
		\multicolumn{10}{|c|}{{5000 training samples; 40\% Query, 30\% Evidence }} \\
		\hline 
		
		BN & -0.274 & -0.721 & -0.436 & -0.565 & -0.675 & -0.474 & -0.229 & -0.509 & -1.223  \\
		DN & -0.268 & -0.634 & -0.317 & -0.499 & -0.610 & -0.407 & -0.185 & -0.490 & -1.089 \\
		SPN & -0.262 & -0.63 & {\bf -0.277} & -0.476 & -0.644 & -0.415 & {\bf -0.118} & -0.489 & {\bf -0.941} \\
		MT & -0.262 & -0.707 & -0.343 & -0.557 & -0.647 & -0.435 & -0.121 & -0.507 & -1.241  \\
		LDFM & {\bf -0.258} &  -0.609 & -0.293 & {\bf -0.460} & {\bf -0.605} &  {\bf -0.399} & -0.166 & -0.481 & -0.991  \\
		LDFM-S & -0.263 & {\bf -0.607} & {\bf -0.291} & -0.462 & -0.613 & -0.462 & -0.130 & {\bf -0.480} & -0.987  \\
		
		\hline 
		
		\hline
		
		\multicolumn{10}{|c|}{{5000 training samples; 30\% Query, 40\% Evidence }} \\
		\hline 
		
		BN & -0.266 & -0.711 & -0.411 & -0.589 & -0.655 & -0.437 & -0.187 & -0.497 & -1.088 \\
		DN & -0.237 & -0.610 & -0.303 & -0.528 & -0.589 &   -0.391 & -0.148 & -0.479 & -0.985 \\
		SPN & -0.229 & -0.619 & {\bf -0.272} & -0.506 & -0.620 & -0.402 & {\bf -0.114} & -0.481 & {\bf -0.893} \\
		MT & -0.226 & -0.698 & -0.348 & -0.603 & -0.620 & -0.427 & -0.116 & -0.499 & -1.188  \\
		LDFM & {\bf -0.230} &  -0.609 & -0.288 &  {\bf -0.481} & {\bf -0.581} &  {\bf -0.383} & -0.129 &  -0.461 & -0.908  \\
		LDFM-S & -0.235 & {\bf -0.588} &  -0.286 & -0.482 & -0.586 & -0.461 & -0.124 & {\bf -0.459} & -0.904  \\
		\hline	
		
		
	\end{tabular}
	
\end{table*}
\subsection{The Derivation Details}

We show the details of deriving the probability of generating an assignment $\mathbf{x}$ discussed in the \emph{LDFM} subsection in the main text.

\begin{align*}
p(\mathbf{x}) & = \sum_{\hat{T}} p(\hat{T}) p(\mathbf{x}|\hat{T}) \\
& = \sum_{\hat{T}} p(\hat{T}) \sum_{M} p(\mathbf{x},M,\hat{T}) \\
& = \beta\sum_{\hat{T}}\sum_{M}\prod_{(i,j) \in E_{\hat{T}}} w_{x_j|x_i} \\
& = \beta n! \sum_{T \in T(G_{\mathbf{x}})} \prod_{(i,j) \in E_{T}} w_{x_j|x_i} \\
& = \beta n! Z_{\mathbf{x}} \propto Z_{\mathbf{x}}
\end{align*}
where $\hat{T}$ is the uniformly generated tree structure and $M$ is a mapping from the $n$ variables to the $n$ nodes of the tree structure $\hat{T}$, $\beta$ is the constant value of $p(\hat{T})$. $\beta n!$ is a constant w.r.t. $\mathbf{x}$. Here we have $n!$ because for each spanning tree $T$ of $G_{\mathbf{x}}$, each permutation of the $n$ variables is generated differently (i.e., corresponds to a different $\langle T,M \rangle$ pair).

\subsection{The Evaluation Results of LDFM-S}

We report the results of LDFM-S and LDFM trained on the 5000-sample datasets and evaluated by using Gibbs sampling on two different proportions of dividing the query and evidence variables in Table \ref{ldfms}. It can be seen that LDFM-S has similar performance to LDFM on most datasets, but achieves significantly better results on the Win95pts dataset and significantly worse results on the Water dataset. Therefore, it may depend on the dataset as to whether modeling distributions over tree structures is useful.

\subsection{More Evaluation Results}

In the \emph{Experiments} section in the main text, we report the evaluation results of two proportions of dividing the query and evidence variables (40\% query, 30\% evidence and 30\% query, 20\% evidence). In Table \ref{other results} we report the evaluation results of the other two proportions (30\% query, 40\% evidence and 20\% query, 30\% evidence).

\begin{table*}[t]
	\small
	\caption{ The maximum of CLL and CMLL normalized by the  number of query variables. The bold numbers mark the best performance.}
	\label{other results}
	\centering
	\begin{tabular}{|l|c|c|c|c|c|c|c|c|c|}
		\hline
		Dataset & Asia & Child & Alarm & Insurance  & Sachs& Water  & Win95pts & Hepar2  & Hailfinder \\
		
		\hline
		\hline
		
		\multicolumn{10}{|c|}{{5000 training samples; 30\% Query, 40\% Evidence }} \\
		\hline 
		
		BN & -0.266 & -0.711 & -0.411 & -0.589 & -0.655 & -0.437 & -0.187 & -0.497 & -1.088 \\
		DN & -0.237 & -0.610 & -0.303 & -0.528 & -0.589 &  -0.391 & -0.148 & -0.479 & -0.985 \\
		SPN & -0.229 & -0.619 & {\bf -0.272} & -0.506 & -0.620 & -0.402 & {\bf -0.114} & -0.481 & {\bf -0.893} \\
		MT & -0.226 & -0.698 & -0.348 & -0.603 & -0.620 & -0.427 & -0.116 & -0.499 & -1.188  \\
		g-LDFM & -0.230 & {\bf -0.609} & -0.288 & {\bf -0.481} & {\bf -0.581} & {\bf -0.383}  & -0.129 & {\bf -0.461} & -0.908  \\
		
		t-LDFM & {\bf -0.210} & -0.630 & -0.349 & -0.556 & -0.590 & -0.389 & -0.159 & -0.464 & -1.019 \\
		
		\hline
		\multicolumn{10}{|c|}{2000 training samples; 30\% Query, 40\% Evidence } \\
		\hline 
		
		BN & -0.266 & -0.764 & -0.469 & -0.599 & -0.669 & -0.466 & -0.195 & -0.506 &  -1.099 \\
		DN & -0.242 & -0.626 & -0.312 & -0.550 & -0.597 & -0.404 & -0.154 & -0.497 & -0.999  \\
		SPN & -0.232 & -0.634 & {\bf -0.300} & -0.519 & -0.634 & -0.406 &  {\bf -0.117} & -0.479 &  -0.911 \\
		MT & -0.228 & -0.719 & -0.371 & -0.623 & -0.638 & -0.446 & -0.132 & -0.521 & -1.250 \\
		g-LDFM &  -0.229 & {\bf -0.587} &  -0.303 & {\bf -0.487}  &  -0.581 & {\bf -0.378}  & -0.126  & {\bf -0.462}  & {\bf -0.900} \\
		
		t-LDFM & {\bf -0.218} & -0.640 & -0.392 & -0.555 & {\bf -0.576} & -0.402 & -0.169 &  -0.465 & -1.013 \\
		\hline
		
		\multicolumn{10}{|c|}{500 training samples; 30\% Query, 40\% Evidence } \\
		\hline 				
		BN & -0.288 & -0.776 & -0.500 & -0.690 & -0.721 & -0.480 & -0.236 & -0.534 & -1.314  \\
		DN & -0.240& -0.654 & -0.348 & -0.706 & -0.625 & -0.418 & -0.186 & -0.498 & -1.058  \\
		
		SPN & -0.243 & -0.752 & -0.425 & -0.637 & -0.741 & -0.512 &  {\bf -0.147} & -0.520 &  -1.140\\
		MT & -0.234 & -0.961 & -0.569 & -0.811 & -0.710 & -0.562 & -0.183 & -0.647 & -2.226  \\
		g-LDFM &  -0.238 & {\bf -0.609} & {\bf -0.331} &  {\bf -0.509} &  -0.596 &  {\bf -0.390} & -0.152 & -0.473 &  {\bf -0.947} \\
		
		t-LDFM & {\bf -0.220} & -0.650 &-0.359 & -0.562 & {\bf -0.579} &  -0.398 & -0.167 & {\bf -0.473} & -1.029 \\
		\hline
		\hline
		\multicolumn{10}{|c|}{5000 training samples; 20\% Query, 30\% Evidence } \\
		\hline
		
		BN & -0.217 & -0.724 & -0.432 & -0.585 & -0.698 & -0.448 & -0.217 & -0.505 & -1.164 \\
		DN & -0.198 & -0.655 & -0.316 & -0.524 & -0.626 & -0.411 & -0.174 & -0.487 & -1.068  \\
		SPN & -0.188 & -0.671 & {\bf -0.297} & -0.514 & -0.660 & -0.428 & {\bf -0.133} & -0.491 &  -0.982 \\
		MT & -0.189 & -0.733 & -0.355 & -0.586 & -0.659 & -0.445 & -0.134 & -0.506 & -1.277  \\
		g-LDFM &  -0.192 &  {\bf -0.644} & -0.309 & {\bf -0.480} &  {\bf -0.617} &  {\bf -0.398} & -0.145 & {\bf -0.468} & {\bf -0.972}  \\
		
		
		t-LDFM & {\bf -0.166} & -0.678 &-0.352 & -0.544 &  -0.618 &  -0.404 & -0.168 & -0.471 & -1.068 \\ 
		
		\hline
		\multicolumn{10}{|c|}{2000 training samples; 20\% Query, 30\% Evidence} \\
		\hline
		
		BN & -0.217 & -0.759& -0.486 & -0.600 & -0.697 & -0.464 & -0.214 & -0.510 & -1.159 \\
		DN & -0.194 & -0.659 & -0.330 & -0.547 & -0.638 & -0.418  & -0.177 & -0.492	 & -1.073  \\
		SPN & -0.189 & -0.682 & -0.324 & -0.518 & -0.667 & -0.431 & {\bf -0.137} & -0.492 &  -0.993 \\
		MT & -0.189 & -0.750 & -0.376 & -0.601 & -0.669 & -0.455 & -0.148 & -0.522 & -1.316  \\
		g-LDFM &  -0.191 &  {\bf -0.637} & {\bf -0.319}  &  {\bf -0.483} &  -0.620 &  {\bf -0.402} & -0.143 & {\bf -0.471} & {\bf -0.946} \\
		
		t-LDFM & {\bf -0.169} & -0.687 &-0.394 & -0.551 & {\bf -0.618} & -0.408 & -0.183 &  -0.472 & -1.062 \\ 
		
		\hline
		
		\multicolumn{10}{|c|}{500 training samples; 20\% Query, 30\% Evidence } \\
		\hline
		BN & -0.223 & -0.778 & -0.522 & -0.664 & -0.716 & -0.479 & -0.262 & -0.535 & -1.431\\
		DN & -0.192 & -0.675 & -0.362 & -0.551 & -0.676 & -0.428 & -0.199 & -0.499	 & -1.116  \\
		SPN & -0.192 & -0.795 & -0.444 & -0.627 & -0.765 & -0.517 & {\bf -0.161} & -0.529 & -1.219  \\
		MT & -0.191 & -0.915 & -0.522 & -0.721 & -0.721 & -0.511 & -0.184 & -0.607 & -2.120 \\
		g-LDFM &  -0.198 &  {\bf -0.648} & {\bf -0.335} & {\bf -0.497} &  -0.642 &  {\bf -0.406} & -0.168 & -0.478 & {\bf -0.990} \\
		
		t-LDFM & {\bf -0.172} & -0.690 & -0.368 & -0.551 & {\bf -0.613} &  -0.410 & -0.175 & {\bf -0.476} & -1.075 \\ 
		\hline
	\end{tabular}
\end{table*}

{
	\bibliographystyle{aaai}
	\bibliography{aaai17}
}

\end{document}